\documentclass{article}
\usepackage{ijcai19}

\usepackage{times}
\usepackage{soul}
\usepackage{url}
\usepackage[hidelinks]{hyperref}
\usepackage[utf8]{inputenc}
\usepackage[small]{caption}
\usepackage{graphicx}
\usepackage{amsmath}%
\usepackage{amsfonts}%
\usepackage{amssymb}%
\usepackage{booktabs}
\usepackage{algorithm}
\usepackage{algpseudocode}
\usepackage{xcolor}
\usepackage{xspace}
\usepackage{tabularx}
\usepackage{xfrac}
\usepackage{multirow}
\DeclareMathOperator*{\argmin}{arg\,min}
\DeclareMathOperator*{\argmax}{arg\,max}
\newcommand{\real}{\mathbb{R}}
\newcommand{\feature}{\real^m}
\newcommand{\loss}{\mathcal{L}}

\newcommand{\concept}[1]{\emph{#1}}
\newcommand{\etal}{et~al.\xspace}
\newcommand{\ie}{i.\,e.\xspace}

\title{A Gradient-Based Split Criterion for Highly Accurate and Transparent Model Trees}
\author{
Klaus Broelemann
\And
Gjergji Kasneci
\affiliations
SCHUFA Holding AG, Wiesbaden, Germany
\emails
\{Klaus.Broelemann,Gjergji.Kasneci\}@schufa.de
}
  
\begin{document}

\maketitle
\begin{abstract}
Machine learning algorithms aim at minimizing the number of false decisions and increasing the accuracy of predictions. However, the high predictive power of advanced algorithms comes at the costs of transparency. State-of-the-art methods, such as neural networks and ensemble methods, result in highly complex models with little transparency.

We propose shallow model trees as a way to combine simple and highly transparent predictive models for higher predictive power without losing the transparency of the original models. We present a novel split criterion for model trees that allows for significantly higher predictive power than state-of-the-art model trees while maintaining the same level of simplicity. This novel approach finds split points which allow the underlying simple models to make better predictions on the corresponding data. In addition, we introduce multiple mechanisms to increase the transparency of the resulting trees.
\end{abstract}

\section{Introduction}
With recent advances in Machine Learning and increasing computational power, automated predictive and decision making systems become widely used and influence peoples lives in many aspects. Applications include personal assistance, advertisement, scoring solutions, fraud prevention and recommendation systems.

Predictive models are influencing our daily lives more and more, thus creating an increasing interest in understandable and transparent predictive models. This interest is opposed by the interest in highly accurate systems. Complex systems, such as neural networks or ensemble models, have shown superior predictive power for a wide range of applications and datasets. This high predictive power comes at the costs of transparency:

``The best explanation of a simple model is the model itself; it perfectly represents itself and is easy to understand. For complex models, such as ensemble methods or deep networks, we cannot use the original model as its own best explanation because it is not easy to understand.''~\cite{lundberg2017unified}

As a consequence, the choice of a predictive model for a specific task includes a trade-off between simplicity and transparency on one side and complexity and higher predictive power on the other side. If the task is mainly driven by the quality of the prediction, complex state-of-the-art models are best suited. Yet, many applications require a certain level of transparency. This can be driven by regulatory constraints, but also by the user's wish for transparency. 
A recent study shows that model developers often feel the pressure to use transparent models~\cite{Veale2018}, even if more accurate models are available.
This leads to simple models, such as linear or logistic regression, where more complex models would have much higher predictive power.
In other cases, transparency is not required, but optional. 
An economic pressure towards models with higher accuracy may lead to low transparency in many important applications. For all these reasons, improving the accuracy of simple models can help to preserve transparency and improve predictions.  

In this work, we present shallow model trees as a way to gain high predictive power with a highly transparent model. A model tree is similar to a decision tree. It consists of a set of rules that form a tree structure. Each leaf contains a simple predictive model. Thus, a model tree is a collection of simple models combined with a set of rules to select a model for each input sample. Shallow model trees consist only of few models and rules. Our experiments show that even a model stump, which uses two simple models, can significantly improve the predictive power, when compared to using only a singe simple model.

\begin{figure}[bt]%
\includegraphics[width=\columnwidth]{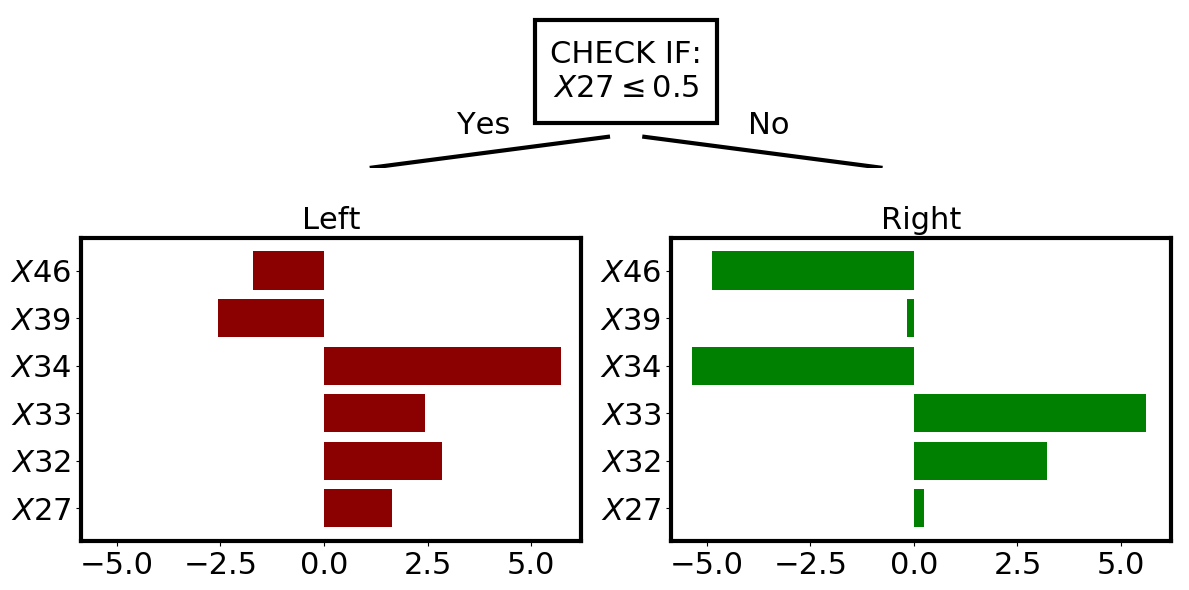}%
\caption{Example explanation of a model stump. The explanation consists of a simple rule in the root node and the influence strength of features in both leaf node models (depicted by the bars). 
This example is taken from the preprocessed Bankruptcy dataset (see evaluation section) and the features such as $X27$ refer to this dataset.
}%
\label{fig:example-explain}%
\end{figure}

Figure~\ref{fig:example-explain} shows a schematic explanation of a model stump. 
The strength of the influence of features is used to explain a simple linear model. The model stump consists of two such models. Both are connected by single rule in the root node.

\subsection{Our Contribution} The main contributions of this work are:
\begin{enumerate}
	\item Shallow model trees as a way to combine transparency with high predictive power.
	\item A novel split criterion for model trees that allows for significantly higher accuracy when compared to state-of-the-art model trees.
	\item The theoretical foundation of the split criterion.
	\item The mechanism of renormalization that further improves predictive power of the presented model trees.
\end{enumerate}

The rest of the paper is organized as follows: In Section~\ref{sec:related-work}, we give a brief overview of related work. Section~\ref{sec:model_trees} introduces our novel method for creating model trees and gives the theoretical background. Section~\ref{sec:renormalization} extends this method to further improve the predictive power. In Section~\ref{sec:evaluation}, we show the effectiveness of our method on various datasets for classification and regression problems. Finally, Section~\ref{sec:conclusion} summarizes this works and gives an outlook and further research.

\section{Related Work}
\label{sec:related-work}
\subsection{Transparency and Explainability}
The recent developments of more and more complex predictive models causes an increasing interest in transparency and explainability of predictions.
Du~\etal present a survey of different explainability approaches~\cite{Du2018}. They distinguish between model-wise and prediction-wise explainability. The former one aims at interpreting a predictive model as a whole, while the latter one explains single preditions. In their comprehensive survey on black box explanation methods, Guidotti~\etal distinguish between four different explanation problems~\cite{guidotti2018survey}: \emph{model explanation} aims at model-wise explanation and \emph{outcome explanation} aims at prediction-wise explanation (analogous to Du~\etal). Additionally, they identify the \emph{black box inspection}, aiming at gaining insides about the unknown interna of a blck box model, and the \emph{transparent design} problem, aiming at finding predictive methods that are explainable by themselves. In this setting, our work mainly addresses the \emph{transparent design} problem. 

Mimic learning is one way to explain a predictive model as a whole. The idea is to approximate a complex model by a simpler, transparent model. Earlier works include approaches that use decision trees as simple models for mimic learning~\cite{Bastani2017,Vandewiele2016,craven1996extracting,schmitz1999ann}. 
Dancey~\etal used logistic model trees for mimic learning~\cite{dancey2007logistic} of neural networks.
Although our algorithm can be used in the same way for mimic learning of complex models, this is out of scope for this work and not covered by our experiments. Another approach for model-wise explainability is to measure the feature importance~\cite{Altmann2010,Hastie2009}.

Prediction-wise explainability does not provide an overall explanation, but concentrates on single samples. One approach towards that goal is to use gradients or Taylor approximations at the given sample point~\cite{Kasneci2016,Bach2015}. Other approaches distort the input to measure the effects of different features on the input~\cite{Fong2017,Robnik-Sikonja2008,Zintgraf2017}

While most of the above work explains existing models, there is also a body of research on explainability by model design. One can argue that mimic learning falls into this category, depending on whether the newly created simple model is used for explanation purposes only or whether it serves as a replacement for the more complex model. Other approaches include constraints to derive simple models~\cite{Freitas2014} and regularized representations~\cite{Zhang2018}. 
By using shallow model trees, this work also deals with explainability by model design. In difference to other approaches, our approach does not add constraints and restrictions to a complex model (except the preference of shallow trees), but rather uses a simple model, which is already highly explainable, and increases its predictive power to make it more attractive for real-world applications.

For a more comprehensive survey on explainability and transparency, we refer to~\cite{guidotti2018survey,Du2018}.

\subsection{Tree-Based Prediction}
Tree-Based models are a widely used class of predictors. A basic form are decision trees~\cite{Safavian1991,Quinlan1993,Hastie2009}, which can be used for both, regression and classification. Decision trees are highly transparent and explainable as long as they are restricted in depth. 

A way to increase the predictive power of decision trees is to use ensembles of trees, such as random forests~\cite{Breiman2001} or gradient boosted trees~\cite{Chen2016}. Again, the increase of predictive power comes at the costs of explainability.

Another class of tree-based predictors are model trees. These are similar to decision trees with the difference that each node contains a simple predictive model instead of a fixed value. The M5 model tree algorithm~\cite{Quinlan1992,Wang1997} for regression is constructed by using decision tree construction rules and is followed by the training of linear models in the nodes. The logistic model tree~\cite{Landwehr2005} warm-starts the training of child-nodes, using the trained model of the parent nodes. 

Another approach for training model trees is to train for each potential split a model in each child node~\cite{potts2005incremental} and to compute for each potential split the loss on the training set. This allows finding the split with respect to the node-models by using the loss as split criterion. On the downside, we see high computational costs that allow to test only a small number of potential split points (such as five per feature dimension in case of \cite{potts2005incremental}).
In difference to existing model tree approaches, our approach does not use existing decision tree constructions methods, but uses a method that is optimized for model trees, which results in improved predictive power. Unlike~\cite{potts2005incremental}, our criterion can be computed much more efficiently and allows testing considerably more split points.

Hierarchical mixtures of experts~\cite{Jordan1994} are a similar concept with probabilistic splits. As a consequence, the final prediction is a probabilistic mixture of all leaf models. Originally, each split depends on a combination of all features, which reduces the explainability. Wang~\etal\cite{Wang2015} present an extension that aims at sparse representations to increases explainability. In difference to our approach, the split criterion depends on multiple features, which results in more complex and less transparent splits. Furthermore, the larger search space in combination with the EM algorithm leads to much higher computational costs. 
Another approach by Sokolovska~\etal~\cite{sokolovska2018provable} avoids hierarchical structures by splitting features independently of each other. This leads to $\prod_{i=1}^m(n_i+1)$ regions / linear models (here, $n_i$ is the number of split points along feature dimension $i$), which is potentially much higher than in tree structures.

\section{Model Tree Construction}
\label{sec:model_trees}
The idea of model trees has already been around for several years.
In this section, we introduce a novel way to construct model trees.
This allows for significantly higher predictive power than state-of-the-art model trees without increasing the number of nodes.

In the following, we briefly introduce current model tree algorithms (in \ref{subsec:mt_algorithm}).
We then introduce the idea of our method (in \ref{subsec:split_idea}). Subsequently, we present the theoretical background of our method (in \ref{subsec:split_theory}). In the end of this section, we present an efficient way to compute the novel split criterion (in \ref{subsec:implementation}).

\subsection{Model Tree Algorithms}
\label{subsec:mt_algorithm}
Model trees have many aspects in common with decision trees, with the major difference that leaf nodes do not contain prediction values, but predictive models instead. Analogously to ensemble methods, we refer to these predictive models as weak models. 

\begin{algorithm}[bt]
\caption{Training Model Trees}
\label{algo:model_tree_training}
\begin{algorithmic}[1]
\Function{BuildNode}{$X\in\real^{n\times m}$, $y\in\real^n$, $d \geq 0$}
\State node$\gets$ NewNode()
\State node.model$\gets$TrainWeakModel($X,y$)
\If {$d > 0$}
	\State $\mathcal{T}\gets$ GetCandidateSplits($X$)
	\State $S\gets\argmax_{S'\in\mathcal{T}} \text{Gain}(X, y, S')$
	\State node.split $\gets S$
	\State node.left $\gets$ BuildNode($X_S$, $y_S$, $d-1$)
	\State node.right $\gets$ BuildNode($X_{\bar{S}}$, $y_{\bar{S}}$, $d-1$)
\EndIf
\EndFunction
\end{algorithmic}
\end{algorithm}

Typically, a model tree is constructed by creating a decision tree and training weak models in the leafs~\cite{Quinlan1992}. 
Algorithm~\ref{algo:model_tree_training} shows a basic version of recursively constructing a model tree. 
Approaches that train one weak model for each split~\cite{potts2005incremental} also fit into the framework of Algorithm~\ref{algo:model_tree_training}, but provide a different (and computationally expensive) way to compute the gain.
We refer to the literature~\cite{Hastie2009} for further details, such as the definition of candidate splits and the gain computation, as well as improvements, such as pruning, smoothing or handling missing data.

Similar to \cite{Landwehr2005}, in the depicted algorithm weak models are trained in all nodes of the tree. This is not required for model trees, but helpful for our novel method.

\subsection{Toward a Better Split Criterion for Model Trees}
\label{subsec:split_idea}

Current model tree construction uses the decision tree construction algorithms with the extension that weak models are trained in the nodes. Model trees typically use the same loss/gain measure as decision trees, e.g. squared error for regression and cross entropy / information gain for classification~\cite{Quinlan1992,Landwehr2005}. 

\begin{figure*}[bt]
	\centering
	\begin{tabular}{ccc}
	\includegraphics[width=0.305\textwidth]{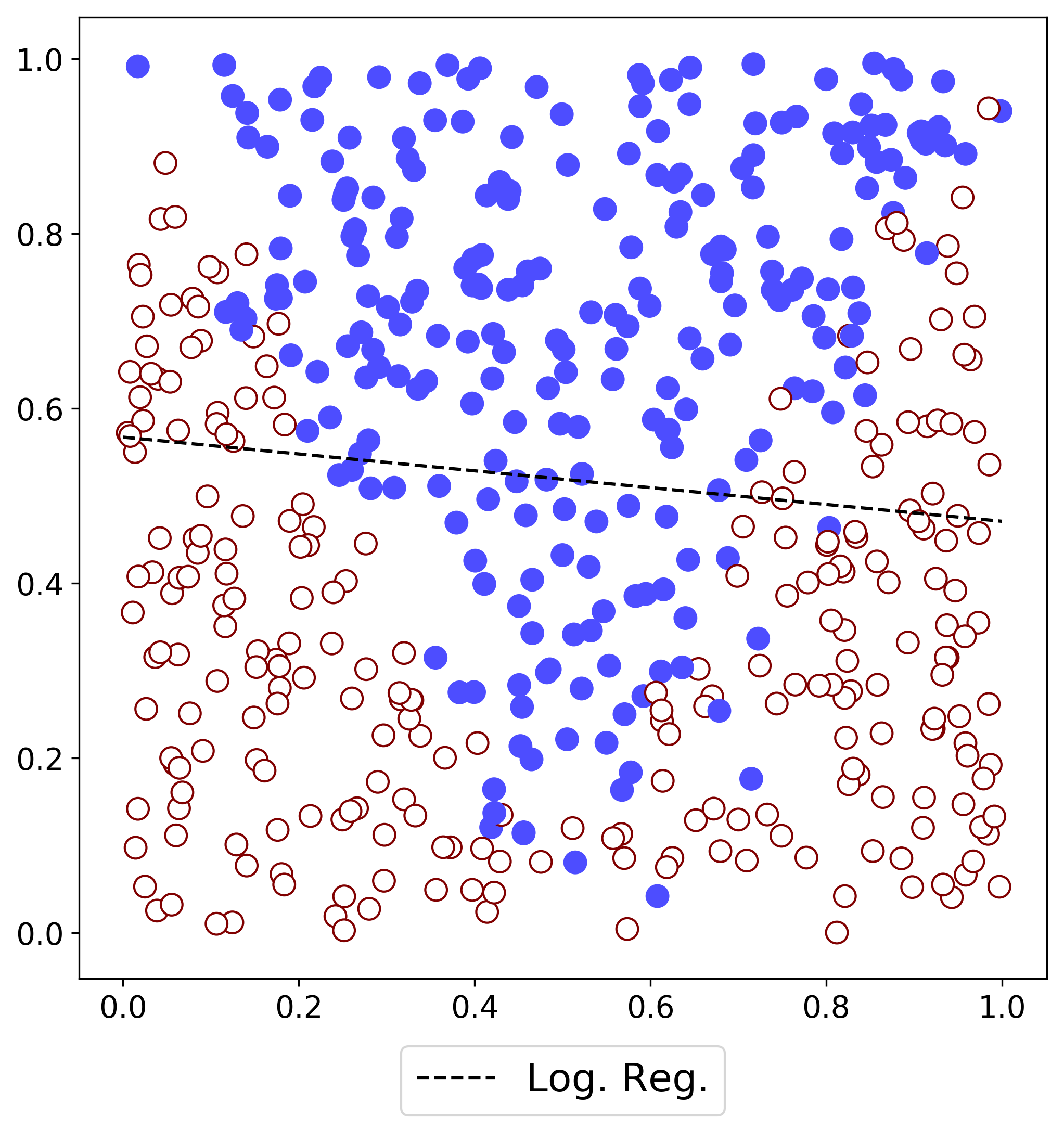}&
	\includegraphics[width=0.305\textwidth]{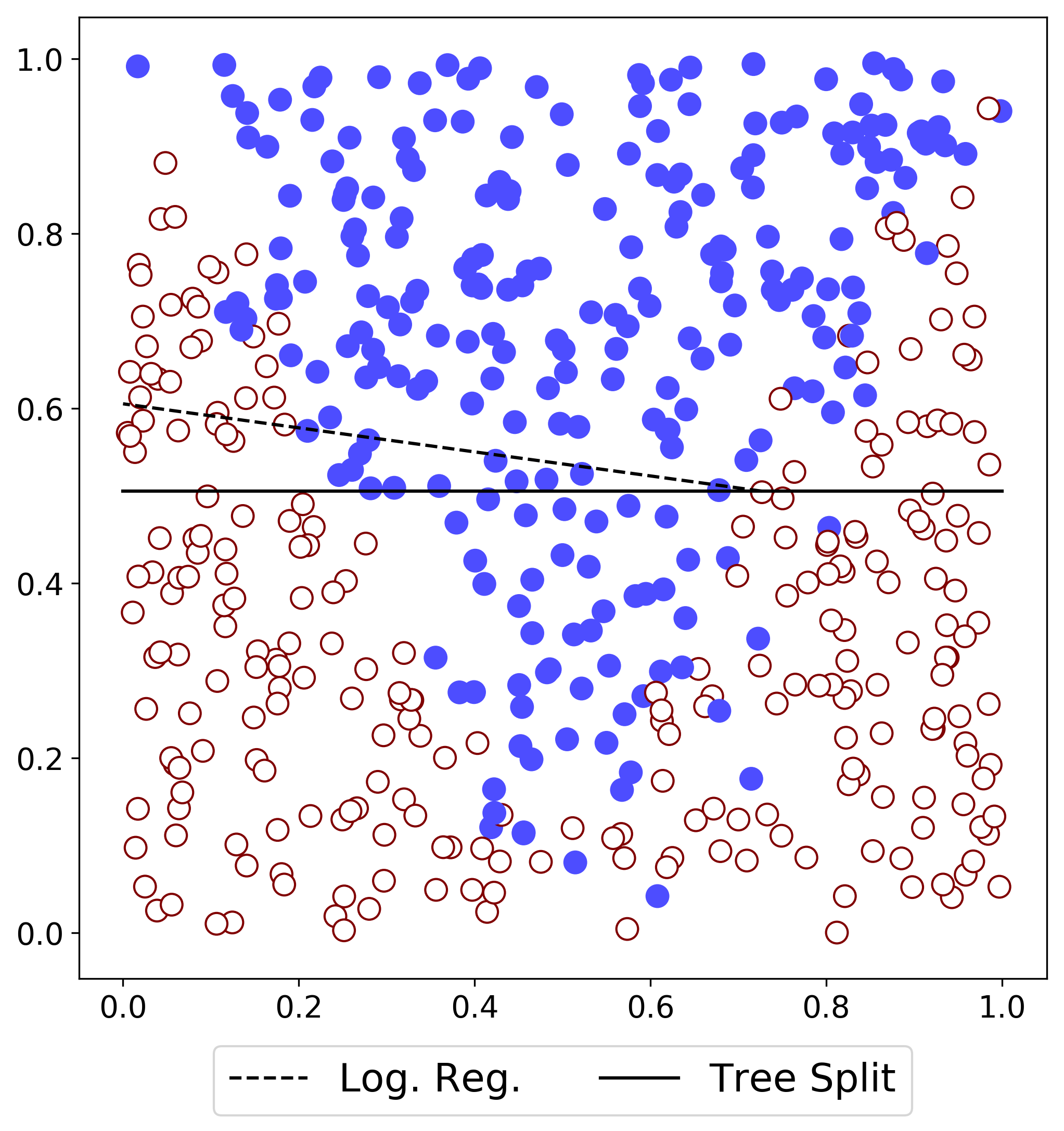}&
	\includegraphics[width=0.305\textwidth]{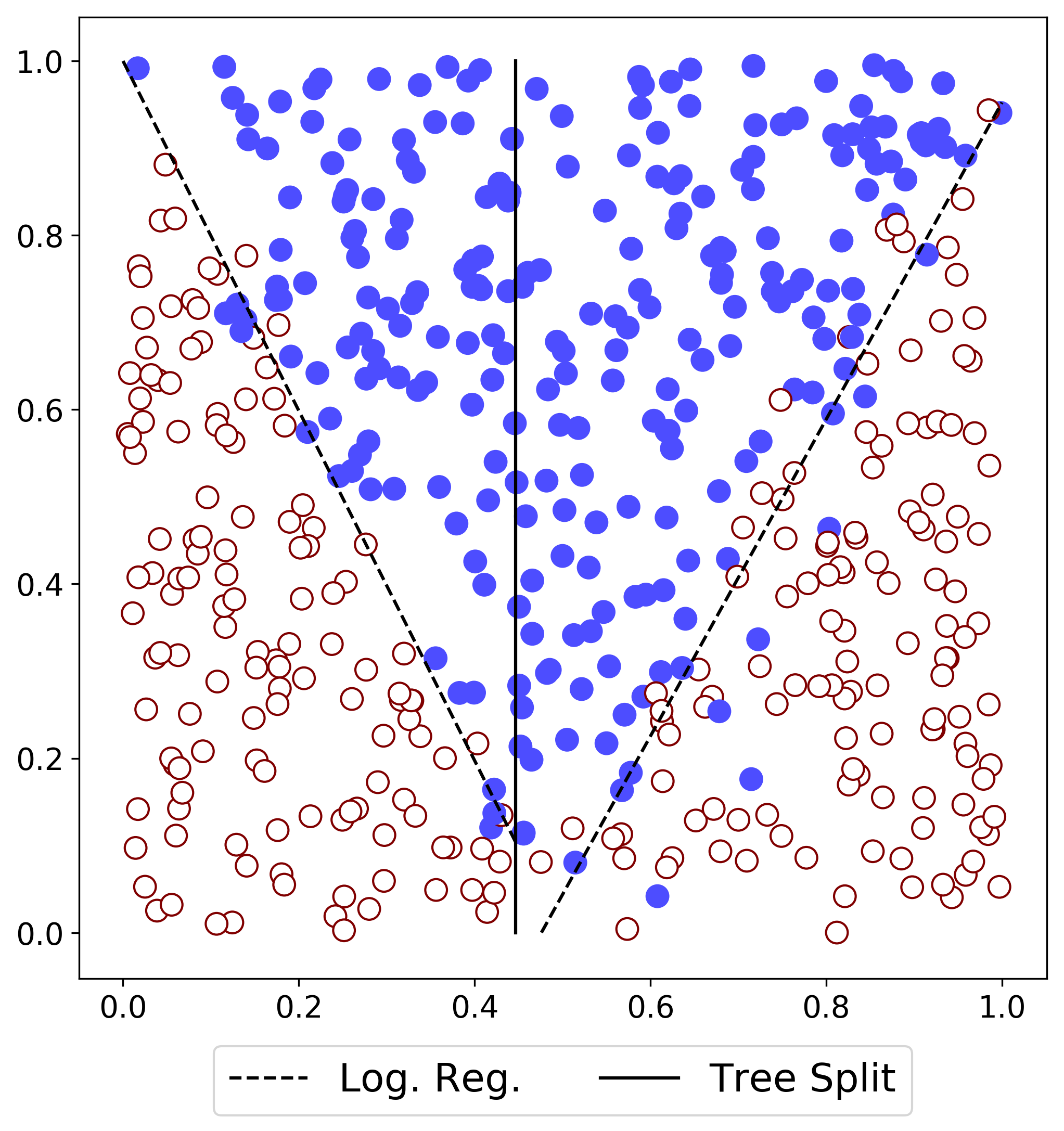}\\
	(a) Logistic Regression & (b) Model tree with cross entropy split & (c) Model tree with gradient based split
	\end{tabular}
	\caption{Comparison of different classification models for an example with two features. (a) Plain logistic regression. The other plots show model trees with one split and logistic regression models in the leafs. Splitting is done (b) by using cross entropy and (c) by using the proposed method. Solid lines are the tree splits. Dotted lines are the 50\% class separators of logistic regression models.}
	\label{fig:example}
\end{figure*}

These loss/gain measures are optimized for decision trees. Although such decision tree measures have been shown to yield good results for model trees, a model-tree-specific gain/loss measure can significantly improve the predictive power of the resulting trees. 
An example can be seen in Fig.~\ref{fig:example}. The cross-entropy measure results in a split that separates both classes in a best-possible way, but gets nearly no improvement from models in the leafs (Fig.~\ref{fig:example}(b)). A better split for model trees would take into account the weak models and makes the classes separable by the weak models instead of directly separating them in the best possible way (Fig.~\ref{fig:example}(c)).

Based on this observation, the major part of this work is a novel gain/loss measure that results in improved model tree predictions without the need to increase the number of nodes. 
Hence, our work improves the predictive power without changing the interpretability of a shallow model trees.

\subsection{Gradient-Based Split Criterion}
\label{subsec:split_theory}
We greedily construct the model tree, similar to Algorithm~\ref{algo:model_tree_training}. 
The major novelty is the way to measure the gain of a potential split of a node.
In the following, we present the theoretical background of our method and derive the split criterion.
A list of symbols and terms used can be found in Tab.~\ref{tab:symbols}.

\begin{table}
\begin{tabularx}{\columnwidth}{lX}
	\toprule
	\bf Symbol & \bf Description\\
	\midrule
	$x_i\in\feature$ & Feature Vector of Sample $i$\\
	$y_i$ & Label of Sample $i$\\
	$I,S\subseteq\{1,\ldots,n\}$ & Index Sets of Samples\\
	$M_\theta:\feature\to\real$ & Weak Model\\
	$\hat{y}_i^\theta=M_\theta(x_i)$ & Prediction of $M_\theta$ for sample $i$\\
	$\loss:\real\times\real\to\real$ & Loss function\\
	$\theta_S$ & Parameters minimizing the loss on $S$\\
	$G_S$ & Gain of a split into $S$ and $\bar{S} = I\backslash S$.\\
	$d$ & Maximal depth of a (sub-)tree\\
	\bottomrule
\end{tabularx}
\caption{List of Symbols}
\label{tab:symbols}
\end{table}

\subsubsection{Training a Weak Model}
Let $M_\theta$ be the class of \concept{weak models} with parameters $\theta$. Our approach handles any weak model that can be trained by (stochastic) \concept{gradient descent}. Nevertheless, with transparency in mind, we recommend to use simple models such as linear or logistic regression.

Given a sequence $(x_i,y_i)_{i=1,\ldots,n}$ of training samples, a loss function $\loss:\real\times\real\to\real$ and a set $I\subseteq\{1,\ldots,n\}$ of samples that correspond to a certain node\footnote{For the root node $I=\{1,\ldots,n\}$. Each child node only holds a subset of the parent nodes training data.}, then training a weak model results in $M_{\theta_I}$ with
\begin{align}
\label{eqn:loss_def}
	\loss_\theta(I) &=\sum\nolimits_{i\in I}\loss\left(y_i, M_\theta(x_i)\right)\\
\label{eqn:optimal_param}
	\theta_I &= \argmin_\theta \loss_\theta(I)
\end{align}

\subsubsection{Optimal Split}
We represent a split of a node by the subset $S\subset I$ of samples that go into the left child node ($\bar{S} = I \backslash S$ is the complement that goes into the right child node). Generalizing from decision trees, the gain of the split $S$ can then be defined as the reduction of the loss by using weak models in the child nodes instead of using one model in the parent node:
\begin{equation}
	G_S = \loss_{\theta_I}(I) - \loss_{\theta_S}(S) - \loss_{\theta_{\bar{S}}}(\bar{S})
\label{eqn:gain}
\end{equation}
Using this gain, the optimal split can be found by computing the gain for each possible split and taking the split with the maximal gain. This requires computing $\theta_S$ and $\theta_{\bar{S}}$ that is training weak models for all possible child nodes. 
This is computationally intractable to do for all potential split points. In \cite{potts2005incremental}, this is done by limiting the number of candidate-splits to five per feature dimension.
To overcome this limitation, we present in the following a method to approximate the gain.

\subsubsection{Gain Approximation}
When optimizing for $\theta_S$, one can warm-start the gradient descent by using $\theta_I$ as an initial value \cite{Landwehr2005}. 
This  will reduce the number of required gradient descent steps and speed up the learning. 
A rough estimation of $\theta_S$ is then already given by a single gradient descent step:
\begin{equation}
	\theta_S \approx \theta_I - \lambda\nabla_{\theta_I}\frac{1}{|S|}\loss_{\theta_I}(S)
\label{eqn:approx_param}
\end{equation}
This estimation works better for simple models, which typically require less gradient descent steps then complex models.

This approximation can be computed efficiently (as we will describe in \ref{subsec:implementation}), but we still have to apply for each set $S$ the model $M_{\theta_S}$ to all samples in $S$ in order to compute the loss $L_{\theta_S}(S)$. Doing so would still be computationally intractable. For this reason, we simplify the computation of the loss using a Taylor approximation around $\theta_I$:
\begin{align}
\nonumber
	\loss_\theta(S) & \approx \loss_{\theta_I}(S) + \left(\theta -\theta_I\right)^T\cdot\nabla_{\theta_I}\loss_{\theta_I}(S)\\
\label{eqn:approx_loss}
	\stackrel{(\ref{eqn:approx_param})}\Rightarrow 
	\loss_{\theta_s}(S) & \approx \loss_{\theta_I}(S) -\frac{\lambda}{|S|}\left\|\nabla_{\theta_I}\loss_{\theta_I}(S)\right\|_2^2
\end{align}
Finally, we get an approximation of the loss:
\begin{align}
\nonumber
	\stackrel{(\ref{eqn:gain})}\Rightarrow
	G_S & \approx \loss_{\theta_I}(I) 
		- \loss_{\theta_I}(S) +\frac{\lambda}{|S|}\left\|\nabla_{\theta_I}\loss_{\theta_I}(S)\right\|_2^2 \\
	\nonumber
		&- \loss_{\theta_I}(\bar{S}) + \frac{\lambda}{|\bar{S}|}\left\|\nabla_{\theta_I}\loss_{\theta_I}(\bar{S})\right\|_2^2\\
\label{eqn:approx_gain}
	&\stackrel{(\ref{eqn:loss_def})}= \frac{\lambda}{|S|}\left\|\nabla_{\theta_I}\loss_{\theta_I}(S)\right\|_2^2 + \frac{\lambda}{|\bar{S}|}\left\|\nabla_{\theta_I}\loss_{\theta_I}(\bar{S})\right\|_2^2
\end{align}
Using this approximation, the gain of all possible split points can be computed and the split with the maximal approximated gain can be used in the tree construction. Here, $\lambda$ is just a scaling factor and can w.l.o.g. be set to $1$. In  the following, we will show an efficient way to this approximation.

\subsection{Implementation}
\label{subsec:implementation}
The split measure (\ref{eqn:approx_gain}) requires to compute the L2 norm of the gradients in both split sets $S$ and $\bar{S}$. This has to be done for every candidate split $S$. Although feasible, the computation is expensive.
The gain-computation for all splits is based on the same sample-wise gradients
\begin{equation}
	g_i = \nabla_{\theta_I}\loss\left(y_i, M_{\theta_I}(x_i)\right)
\label{eqn:sample-wise_gradient}
\end{equation}
These only need to be computed once per node. For continuous features, every split candidate is given by a feature index $j$ and a threshold $t$, which leads to $S = \{i\in I|x_{i,j}\leq t\}$. By sorting along values of feature $j$, the split measure (\ref{eqn:approx_gain}) can efficiently be updated. Algorithm~\ref{algo:gradient_model_tree_training} shows an efficient way to compute the split criterion by updating the sum of the gradients. 
This can then be used as an implementation for lines 5 and 6 in Algorithm~\ref{algo:model_tree_training}.

This can even be improved further by using the fact that $\theta_I$ is optimized on $I$. This means that the gradient $g_I$ is approximately zero (otherwise, $\theta_I$ is not optimized). As a consequence, $g_{\bar{S}} = -g_S$.
Due to practical considerations, such as stopping criteria for stopping before reaching $g_I = 0$ or the usage of mini-batches, we do not directly set $g_I$ to zero, but compute it as described in line 3 of Algorithm~\ref{algo:gradient_model_tree_training}.

Note that our algorithm trains {\bf exactly one model} for each node. It then uses gradients $g_i$ of this trained model to approximate the gain and to find the optimal split point. With this split point, we train afterwards exactly one weak model for each child node.
The training of child nodes can be warm-started by the learned parameters $\theta_I$ of the parent node's weak model.

\begin{algorithm}[bt]
\caption{Gradient-Based Split Finding}
\label{algo:gradient_model_tree_training}
\begin{algorithmic}[1]
\Require Index Set $I$, Optimized Parameters $\theta_I$, Training Data $(x_i,y_i)_{i\in I}$
\State $j_{Opt},t_{Opt},gain_{Opt}\gets -1$
\State $g_i\gets\nabla_{\theta_I}\loss\left(y_i, M_{\theta_I}(x_i)\right)$
\State $g_{I}\gets\sum_{i\in I}g_i$
\State $g_{S}\gets (0,0,\ldots,0)^T$
\For{feature $j=1,\ldots,m$}
	\State Sort $g_i$ and $x_i$ according to feature $x_{\cdot,j}$
	\For{Split Point $t_k$ along feature $j$}
		\State $U\gets \{i\in I|t_{k-1} < x_{i,j} \leq  t_k\}$
		\State $g_{S} \gets g_{S} + \sum_{i\in U} g_i$
		\State $g_{\bar{S}} \gets g_{I} - g_{S}$
		\State $gain\gets \left\|g_{S}\right\|_2^2 + \left\|g_{\bar{S}}\right\|_2^2$
		\If{$gain > gain_{Opt}$}
			\State $j_{opt}\gets j$; $t_{opt}\gets t_k$; $gain_{opt}\gets gain$
		\EndIf
	\EndFor
\EndFor
\State \Return{$j_{Opt},t_{Opt},gain_{Opt}$}
\end{algorithmic}
\end{algorithm}                                       

\section{Renormalization}
\label{sec:renormalization}
The previous section introduced the gradient-based split criterion. Due to the approximations used, the split quality measure directly depends on the gradient. Unfortunately, the gradients are often not translation invariant, \ie shifting the feature values can change the optimal split point (in terms of criterion (\ref{eqn:approx_param})). Furthermore, gradient descent often shows low performance and slow convergence when the input data is poorly scaled. Note that each split shrinks / restricts the domain of one feature for the child nodes. As a consequence, data that was normalized on the whole training set might lead to poorly normalized subsets in the leafs.

We renormalize the training data to work with normalized data in each node. 
We then get an improved, translation invariant version of approximation~(\ref{eqn:approx_param}) and also speed up and improve the model training via gradient descent by providing well scaled input features.
The renormalization only takes into account the training data from set $S$. As a consequence, the renormalization is done for each potential split individually. 

A basic approach is to renormalize the features for $S$ and $\bar{S}$, compute the gradients on these normalized features and compute the loss according to (\ref{eqn:approx_loss}). Computing all gradients anew for each potential split is intractable. In addition, the weak model of the parent node is trained on differently normalized features, which complicates the gradient computation. 

In the following, we present the mathematical background of our renormalization approach. 
The theory leads to a computationally more efficient way to compute the gradients from renormalized features. 
Subsequently, we present the explicit computation for linear models, such as linear and logistic regression, in combination with linear renormalizations, such as z-normalization and min-max-normalization.

\subsection{Theory} Given a potential split $S\subset I$, the corresponding training features $x_i$ can be normalized, e.g. by using a z-normalization. In the most general form, the normalization is a transformation $T_S:x_i\mapsto \tilde{x}_i$ from original feature space into a normalized feature space. We further restrict $T_S$ to be compatible with the model, i.e. for each parameter $\theta$, there must exist a parameter $\tilde\theta$ with
\begin{equation}
	M_\theta(x_i) = M_{\tilde\theta}(\tilde{x}_i)
	\label{eqn:norm_constraint}
\end{equation}
As a consequence, the transformed loss $\tilde\loss$ can be written as:
\begin{equation}
	\tilde\loss_{\tilde\theta}(S) := \sum\nolimits_{i\in S}\loss(y_i, M_{\tilde\theta}(\tilde{x}_i))\\
				= \loss_\theta(S)
\label{eqn:loss_def_normalized}
\end{equation}
Analogously to (\ref{eqn:approx_loss}), the loss can be approximated using gradients in the normalized feature space:
\begin{align}
\nonumber
	\loss_{\theta_S}(S) = \tilde\loss_{\tilde\theta_S}(S) 
	&\approx \tilde\loss_{\tilde\theta_I}(S) - \frac{\lambda}{|S|}\left\|\nabla_{\tilde\theta_I}\tilde\loss_{\tilde\theta_I}(S)\right\|_2^2\\
	&\stackrel{(\ref{eqn:loss_def_normalized})}= \loss_{\theta_I}(S) - \frac{\lambda}{|S|}\left\|\nabla_{\tilde\theta_I}\loss_{\theta_I}(S)\right\|_2^2
	\label{eqn:approx_loss_normalized}
\end{align}
Note that there is a functional dependency between $\theta$ and $\tilde{\theta}$, which allows computing the last gradient in (\ref{eqn:approx_loss_normalized}). 
For simple models, such as linear or logistic regression, Equation (\ref{eqn:approx_loss_normalized}) often gives better approximations of the gain than (\ref{eqn:approx_loss}). For such models, normalizing the features speeds up the gradient descent convergence and leads to better approximations after one step.

Still, normalizing and computing the gradients for each step individually is computationally infeasible. Rewriting the gradients can help to find a faster computation:
\begin{align}
	\nabla_{\tilde\theta_I}\loss_{\theta_I}(S) &=\left(\nabla_{\theta_I}\loss_{\theta_I}(S)\right)^T \cdot \frac{\partial \theta_I}{\partial \tilde\theta_I}
\end{align}

This means that only one computation of the gradient per node is required. It can then be corrected using the derivative $\frac{\partial \theta_I}{\partial \tilde\theta_I}$. This can be done in Algorithm \ref{algo:gradient_model_tree_training} in line 11 by adding this sample-independent factor.
Note that this derivative has to be computed for each potential split set anew.
In the linear case, this can be done efficiently:

\subsection{Linear Case} Linear models have the form
\begin{equation}
M_\theta(x) = f(w^Tx + b),\qquad \theta = (w,b)
\label{eqn:lin_model}
\end{equation}
with a function $f:\real\to\real$. For linear regression, $f$ is the identity, while for logistic regression $f$ is the logistic function.
Linear normalizations can be written as 
\begin{equation}
	\tilde{x} = T_S(x) = Ax + c
\label{eqn:lin_normalization}
\end{equation}
where $A$ is a diagonal matrix, e.g. for z-normalization, $A$ contains the inverse standard deviations of all features.
Based on (\ref{eqn:norm_constraint}) and (\ref{eqn:lin_model}) follows:
\begin{align*}
	x &= A^{-1}\tilde{x} - A^{-1}c & 
	w &= A\tilde{w} & b &= \tilde{b} + \tilde{w}^Tc
\end{align*}
The partial derivative can then be written as:
\begin{equation}
	\frac{\partial \theta_I}{\partial \tilde\theta_I} = \left(\begin{array}{cc}A & 0\\ c^T & 1\end{array}\right)
\label{eqn:partial_theta}
\end{equation}
For z-normalization, $A$ and $c$ depend on mean and the standard deviation. For  min-max normalization, $A$ and $c$ depend on minimum and maximum. Both cases can efficiently be updated when computing the sum of the gradients (see also Algorithm~\ref{algo:gradient_model_tree_training} for updating computations). 

It is worth noting that once a model with linear weak models is trained, one can create an equivalent model that operates solely on the unnormalized data by applying the inverse normalization steps on split points and model weights. This can be used speed up the predictions done with such a model, e.g. in real-time scenarios where each prediction has to be performed within few milliseconds.

\subsection{Example} 
\begin{figure}[bt]%
\includegraphics[width=\columnwidth]{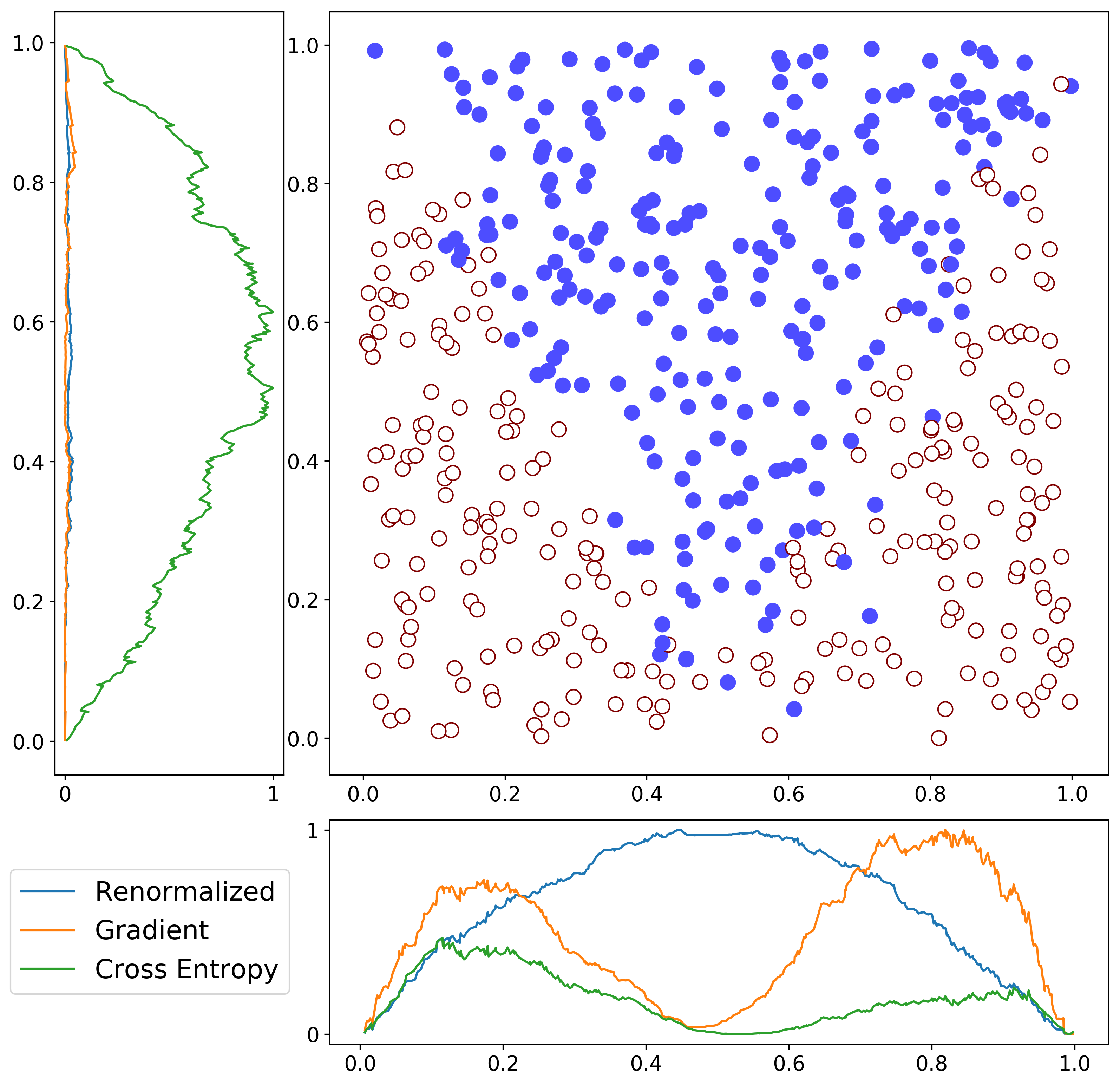}%
\caption{Different gain measures along the two axes. The gain-plots are normalized so that the maximal gain (across both feature dimensions) of a measure (\ie. split measure value at the chosen split point) gets a value of 1. One can see, that the cross entropy measure leads to a horizontal split, while both gradient-base measures have the maximum at the first axis (\ie a vertical split) with the renormalized version finding the best split for this example.}%
\label{fig:example_gain}%
\end{figure}

In order to demonstrate the improvement from renormalizing the data, we computed the split gain along the two axes of the previous example. Figure~\ref{fig:example_gain} shows three different methods to compute the gain. It can be seen that cross entropy finds the best split on the y-axis while the presented gradient-based method finds a better point on the x-axis. Still, the ideal point (in terms of the loss of the resulting model tree) can be found using the renormalization technique.

\begin{table*}[!bt]%
\centering
\begin{tabular}{l|l||l|l|l||lll|lll|lll}
\toprule
& \bf Method & {\bf Linear} & {\bf DT} &   {\bf XGB} & \multicolumn{3}{c|}{{\bf MT-G}} & \multicolumn{3}{c|}{{\bf MT-GR}} & \multicolumn{3}{c}{{\bf MT-DT}} \\
& \bf Depth &       \empty &   \empty &      \empty &          1 &     2 &     3 &           1 &     2 &           3 &            1 &     2 &     3 \\
\midrule
\multirow{5}{*}{(a)}&Bankruptcy    &         80.8 &     85.0 &  {\bf 93.2} &        88.6 &  91.1 &  {\bf 91.2} &        88.6 &        90.4 &        91.2 &       82.2 &  87.8 &  90.7 \\
& Breast Cancer &         99.4 &     92.6 &        99.2 &        99.6 &  99.1 &        99.1 &        99.6 &  {\bf 99.7} &        99.4 &       99.0 &  98.2 &  98.2 \\
& Census        &   {\bf 94.1} &     89.6 &  {\bf 94.7} &  {\bf 94.0} &  93.8 &        93.7 &        93.9 &        93.7 &        93.3 &       93.9 &  93.9 &  93.7 \\
& Credit Card   &         71.7 &     73.8 &  {\bf 76.2} &        71.7 &  74.9 &        74.8 &        74.4 &        75.3 &  {\bf 75.6} &       72.1 &  74.8 &  75.1 \\
& Olympics      &         75.2 &     63.8 &        75.8 &        76.1 &  76.6 &        76.8 &        76.2 &        77.3 &  {\bf 78.3} &       76.2 &  76.3 &  77.1 \\
\midrule
\multirow{3}{*}{(b)}&House        &         77.2 &     78.3 &        87.9 &       82.2 &  83.4 &  83.8 &        83.9 &  86.3 &  {\bf 88.2} &         77.9 &  82.6 &  83.7 \\
& Weather-Temp &         62.1 &     85.7 &  {\bf 90.1} &       69.7 &  78.6 &  84.0 &        80.4 &  87.3 &  {\bf 88.3} &         64.9 &  70.8 &  79.4 \\
& Weather-Vis  &         24.2 &     57.9 &  {\bf 63.9} &       43.2 &  49.7 &  55.5 &        46.2 &  54.5 &  {\bf 58.0} &         43.6 &  51.9 &  56.0 \\
\bottomrule
\end{tabular}
\caption{Evaluation results for (a) classification and (b) regression. Highlighted are results that are equaly to or better than the  best model trees result.}
\label{tab:results-c}
\end{table*}

\section{Evaluation}
\label{sec:evaluation}
The previous sections introduced our novel method to create model trees. In this section, we will evaluate the effectiveness of this novel method on multiple datasets. The experiments cover both classification and regression.

\subsection{Datasets}
\urldef\censusurl\url{https://archive.ics.uci.edu/ml/datasets/Census-Income+%28KDD%29}
For the experiments, we used the multiple datasets. These datasets include financial and medical datasets, where decisions can have a high impact and, hence, transparent predictions are required. All datasets come from public sources\footnote{
\url{https://archive.ics.uci.edu/ml/datasets/Polish+companies+bankruptcy+data}, 
\url{https://archive.ics.uci.edu/ml/datasets/Breast+Cancer+Wisconsin+(Diagnostic)},
\censusurl,
\url{https://archive.ics.uci.edu/ml/datasets/default+of+credit+card+clients},
\url{https://www.kaggle.com/heesoo37/120-years-of-olympic-history-athletes-and-results},
\url{https://www.kaggle.com/harlfoxem/housesalesprediction},
\url{https://www.kaggle.com/budincsevity/szeged-weather}
}~\cite{Yeh2009,Zikeba2016} and cover both classification and regression tasks.
The number of samples and attributes of these datasets are displayed in Tab.~\ref{tab:datasets}.

\begin{table}[bt]%
\begin{tabular}{lrrlr}
	\toprule
	\bf Dataset & \bf \#Samples & \multicolumn{2}{r}{\bf \#Attr. [Col.]} & \bf Preval.\\
	\midrule
	Bankruptcy & 43,405 & 64 && 4.82\% \\
Breast Cancer & 569 & 30 && 37.26\% \\
Census & 299,285 & 507 && 6.20\% \\
Credit Card & 30,000 & 24 & [33] & 22.12\% \\
Olympics & 271,116 & 8 & [304] & 14.67\% \\

	\midrule
	House & 21,613 & 19 \\
Weather-Temp & 96,453 & 10 & [11] \\
Weather-Vis & 96,453 & 11 & [12] \\

	\bottomrule
\end{tabular}
\caption{List of datasets. For each dataset the number of samples and attributes is listed. A second number for attributes indicates the number of columns after one-hot-encoding categorical attributes. For classification datasets the prevalence of class 1 is also denoted.}
\label{tab:datasets}
\end{table}

\subsection{Methods}
We compare the performance of different methods with each other. As a baseline, we use the weak models Lineare and Logistic Regression (Lin) as well as Decision Trees (DT). As a reference complex model, we use XGBoost (XGB)~\cite{Chen2016}. This model has been shown to be robust towards the choice of hyper parameters.
Furthermore, we use model trees with state-of-the-art split criteria od decision trees (MT-DT). These are mean-square-error for regression and the entropy-based information gain for classification. 

All these methods are compared with two versions of our model trees. The original version from Section~\ref{sec:model_trees} (MT-G) and the version using renormalization (MT-GR). With transparency and shallow trees in mind, we restrict all model trees to a fixed depth of 1, 2, or 3. In our experiments, we evaluate the effect of using different maximal depth limits.

\subsection{Experiments}
For our experiments, we performed 4-fold cross validation and averaged the 4 performance measurements. 

\paragraph{Classification.} We use the area under the roc-curve (auc) as measure for classification. The classification results are shown in Tab.~\ref{tab:results-c} (a).
It can be seen that the gradient-based model trees with renormalization provide the best results, often even a model stump (depth $=1$) results in significantly higher auc than logistic regression. The exception is the census dataset that can sufficiently be modeled by logistic regression. Additional splits in models trees only create slight overfitting effects.
In two cases, model trees even out-perform XGBoost.

\paragraph{Regression.} We evaluate the regression experiments using the $r^2$-measure. The results are shown in Tab.~\ref{tab:results-c} (b). Our results show that the renormalized version of our splitting criterion consistently creates the best results and each new level of depth increases the predictive power.

\section{Conclusion}
\label{sec:conclusion}
Starting with the motivation to find a transparent model with high predictive power, our work concentrated on shallow model trees. While state-of-the-art model trees are highly transparent, their predictive power bears potential for improvement. In this work, we presented a novel way to create model trees, which has a strong theoretical foundation, is efficient and results in higher predictive power than current model tree algorithms.
Given a high requirement of transparency, our gradient-based model trees are powerful method for automated predictive and decision making systems.

Based on this work, we aim at further improvements of this method. This includes ways to enhance the explainability, e.g. through regularization terms that prefer round numbers as split points. 

\bibliography{references}
\bibliographystyle{named}
\end{document}